# Comparison of Varied 2D Mapping Approaches by Using Practice-Oriented Evaluation Criteria

Vergleich verschiedener 2D Kartenerstellungsverfahren unter Verwendung Praxis-Orientierter Evaluationskriterien

*Justin Ziegenbein*
*Manuel Schrick*
*Marko Thiel*
*Johannes Hinckeldeyn*
*Jochen Kreutzfeldt*

*Institute for Technical Logistics*
*Hamburg University of Technology (TUHH)*

**A** key aspect of the precision of a mobile robot's localization is the quality and aptness of the map it is using. A variety of mapping approaches are available that can be employed to create such maps – with varying degrees of effort, hardware requirements and quality of the resulting maps. To create a better understanding of the applicability of these different approaches to specific applications, this paper evaluates and compares three different mapping approaches based on simultaneous localization and mapping, terrestrial laser scanning as well as publicly accessible building contours.

*[Keywords: mobile robots, mapping, map evaluation, localization]*

**E**in wichtiger Aspekt für die präzise Lokalisierung von mobilen Robotern ist die Qualität und Eignung der verwendeten Karte. Eine Vielzahl von Ansätzen stehen zur Erzeugung solcher Karten zur Verfügung, die in Punkten wie Aufwand, benötigter Hardware und nicht zuletzt der Qualität ihrer Ergebnisse teils stark variieren. Um ein besseres Verständnis von diesen Ansätzen und deren Eignung für konkrete Anwendung zu schaffen, werden in dieser Arbeit drei verschiedene Ansätze evaluiert und miteinander verglichen: SLAM, Terrestrischer Laser Scanner und öffentlich verfügbare Gebäudekonturen.

*[Schlüsselwörter:Mobile Roboter, Kartenerstellung, Kartenevaluation, Lokalisierung]*

## 1 INTRODUCTION

Localization is one of the most crucial tasks for mobile robots. Being able to determine the robot's location is vital for safe navigation and thus the success of the overall process. For this, robots typically use a 2D map that shows the contours of the respective area visible to the robot. The robot then matches its sensor data with these contours to compute its position. An increased resemblance to reality and higher levels of detail lead to a higher level of precision in the robot's localization.

Multiple approaches to creating such maps have been presented in the literature that show different benefits and drawbacks not only in the resulting maps but also in the effort and technical equipment required in the mapping process. The ones employing simultaneous localization and mapping (SLAM) are the most prominent approaches in the literature [1], [2]. These rely on the robot itself and thus typically yield results with the same level of detail as supported by its sensors. However, a lot of algorithmic effort is put into reducing the amount of inaccuracy and noise created by the hardware and its movements. An extensive, robot-specific setup and parametrization phase are necessary to endure the validity and quality of the map. Other approaches such as [3] employ stationary terrestrial laser scanners (TLS) to create a map that shows a high level of detail and precision. However, the external hardware introduces additional acquisition costs and the mapping process can be time-consuming and may require extensive postprocessing effort. Lastly, approaches such as [4] make use of publicly accessible building contours (PABC) extracted from satellite data. These approaches require a low amount of effort and no specific hardware in their process. However, they require data of the respective area to be available and generate maps with a low level of detail.

With their individual benefits and drawbacks, none of these approaches is suited or even applicable for every kind of robotic application. To support the decision process, several evaluations and benchmarking methods have been introduced in the literature [5–8]. However, these only concentrate on quantifying the quality of the map and do not consider the effort or specific hardware requirements. It is also unclear to what extent these quantitative measures of map quality correlate to their aptness for a specific application. In addition, current publications only compare the results of different implementations of the same approach. To the best of our knowledge, a cross-approach evaluation of mapping systems has not been published yet.

To address this research gap, this contribution will compare three different mapping approaches, namely SLAM, TLS and PABC. This is done using a novel evaluation approach that combines the quantitative analysis of map quality, effort and hardware requirements with a practical assessment of the constructed maps in a real-world scenario. In addition, the results will be used to examine correlations between the required effort and the resulting map quality.

The contribution is structured as follows: Section 2 gives an overview of map evaluation methods presented in the literature. In Section 3, the map creation process of each of the three mapping approaches is presented, as well as the method that is followed to carry out the evaluation. The selection of the candidate implementation for each mapping approach as well as the area in which the evaluation took place and the hardware that has been used is presented in Section 4. The results are presented and discussed in Section 5. Finally, Section 6 concludes this paper and states potential for future work.

## 2 MAPPING APPROACH EVALUATION METHODS

The methods to evaluate mapping approaches presented in the literature can be categorized into trajectory-based, procedure-based and map-based evaluation methods. All of them aim to provide a consistent and reproducible evaluation.

Trajectory-based evaluation methods compare the trajectory provided by a localization algorithm to a ground truth trajectory. It is mainly used to evaluate the localization performance of SLAM approaches with one another and evaluates only implicitly the quality of the map. This type of evaluation often makes use of open-access benchmark datasets. They provide different sensor data (e.g. LiDAR, IMU, camera) as well as ground truth monitoring of the trajectory in suitable surroundings. Examples of such benchmark datasets are the KITTI-dataset provided by the KIT [9] and DARPA provided by the MIT [10]. Both include sensor data for visual- and laser-SLAM algorithms.

The difference between ground truth and the trajectory built by SLAM is evaluated using different metrics. Sturm et al. use the Absolute Pose Error (APE) for estimating global consistency and the Relative Pose Error (RPE) to measure the local consistency of the provided trajectory over a defined period of time [11]. Burgard et al. manually correct the calculated trajectory as ground-truth. They also use a relative error which is independent of previous localization imprecisions [6, 7]. Dhaoui et al. suggest using the Hausdorff distance to measure the localization error [12]. Most of the approaches require a precise ground truth measure which is expensive in terms of workload. Abdallah et al. therefore use GPS-Data as a reference and referenced air view pictures as map reference [13].

The second category of evaluation methods is procedure-based evaluation. In the literature, this form of evaluation mostly considers SLAM approaches. A well-known procedure for the evaluation of map generation procedures is the theoretical evaluation of selected criteria. These can include parameters used for the calculation of the robot position, the type of map, sensor information used or assumptions on which the SLAM methods are based [14].

Often the measurement of the CPU utilization or the processed frames per second (FPS) as well as memory utilization is evaluated. On the one hand, conclusions can be drawn as to which robot platforms are suitable for the respective process. On the other hand, basic technical requirements that need to be fulfilled in order to use the SLAM algorithm are analyzed [15–17].

The third category of map-based evaluation considers the quality of the end product, meaning the map itself. In order to evaluate the map quality independent of the mapping process and localization performance, additional map-related metrics are presented below. These include the shape of the map, distances between two superimposed maps, artificial and natural features in the map and information content of the map.

The structure-based map evaluation deals with the deviations of maps to a ground truth. Often the average deviation between a reference map and the examined map is calculated. For the quantitative comparison, the ADNN (Average Distance Nearest Neighbor) algorithm can be used [15]. Another way of evaluating the similarity of two maps is the Structural Similarity Index Measure. It is used to measure the optical correspondence of two images [12]. In addition, the ICP (Iterative Closest Point) method is also used to iteratively search for the best match between maps. After a certain number of iterations, the deviation between the maps is calculated in form of the maximum, the average and the minimum error [12].

Another map-based alternative is the feature-based methodology. Here, individual features that are present in both, the reference map and the map under study, are identified and compared. One example for such features are corners in walls and other obstacles. The more the number of corner deviates from other maps, the higher the probability of inconsistencies and artifacts in the map. The Harris Corner Detector is often used to identify corners in maps [18, 19]. An alternative approach to find features in a map is the transformation of the map into Hough space and the subsequent identification of extrema using SIFT (Scale-invariant feature transform) [18, 20]. The proportion of assigned features of the map to the total number in a reference map then serves as a quality measure for evaluation. Another possibility is presented by Chen et al. with LSF (Least Square Fitting), where rotations and translational parameters are first computed to align the reference map. Then the RMSE (Root Mean Square Error) between selected feature points is calculated and compared [5].

Schwertfeger et al. introduce the Fiducial Map Metric as a way to evaluate features independently of a reference map. Artificial features in the form of cylinders separated by a wall are set up. The position of the fiducials in reality in comparison to their corresponding position in the map is used for the evaluation [21]. Further approaches to map evaluation are the optical comparison or the superimposition of maps in order to recognize at which point deviation exists [17, 22].

Existing evaluation approaches largely consider SLAM as mapping method. In some cases, data from a terrestrial laser scanner is used as ground truth or reference map due to its high accuracy. A procedure to compare mapping methods with different approaches has not been published yet. A practice-oriented easily applicable cross-approach evaluation over all categories cannot be found. Therefore, in the following a new method to evaluate mapping procedures and the quality of the map as well as its localization performance is introduced and executed experimentally on three different mapping approaches.

## 3 METHOD

This chapter will introduce the method used in this paper. To do so, the process of mapping in the three different categories SLAM, TLS and PABC will be presented. Afterward, the evaluation method will be displayed.

The SLAM mapping process can be divided into three steps. First, the sensor data is gathered using a mobile robot. For this purpose, a route through the test area is planned with the goal of reaching every area at least twice. The LiDAR, odometry and IMU data is saved in recordings to make the process of mapping reproducible and to allow parameter optimization on the same data set. After selecting a SLAM-Algorithm it is installed on a workstation. This allows the algorithm to be executed without being limited by the robot's computing power. Second, the algorithm is parameterized to according to the robot's hardware setup. Third, the recorded sensor data is used to optimize the parameters of the algorithm iteratively. Once a satisfactory setup is found, the map is exported.

For TLS the process is divided into four steps. First, the test area is examined to plan the scan positions such that it won't create blind spots. Artificial, spherical targets are used for an easier registration of the different scans. These are positioned such that two consecutive scans share at least two visible spheres. Second, the scans are conducted, moving the TLS and the spheres manually to each planned position. Third, the resulting scans are linked with one another in the scan registration. This results in a 3D-Point Cloud which is subsequently cleaned from irrelevant data for the 2D-representation, such as balconies, which are not in the field of view of the robot or the floor. In the last step a top-down 2D-representation of the point cloud is extracted and formatted to be used as a map.

The procedure for using PABC requires three steps. First, a suitable database for the map is obtained. The map requires a high resolution and a high level of detail. Second, the contours of buildings and other important features in the test area are extracted from the database. Third, the building contours are transformed and exported into the required black and white picture format.

To address the identified research gap, a practice-oriented cross-approach evaluation is used that takes all three before mentioned evaluation approaches into consideration and can therefore give an overview of the localization performance, map accuracy and the mapping process. The basis of the evaluation approach is shown in Figure 1. The evaluation procedure is described in the following.

First, the three mapping approaches are conducted in a suitable environment. The time required to create the maps with the three different mapping approaches is recorded for the later calculation of a quality to effort ratio. Specific data to be recorded are defined, and reference measurements are performed. Test scenarios are used to evaluate the map generation methods.

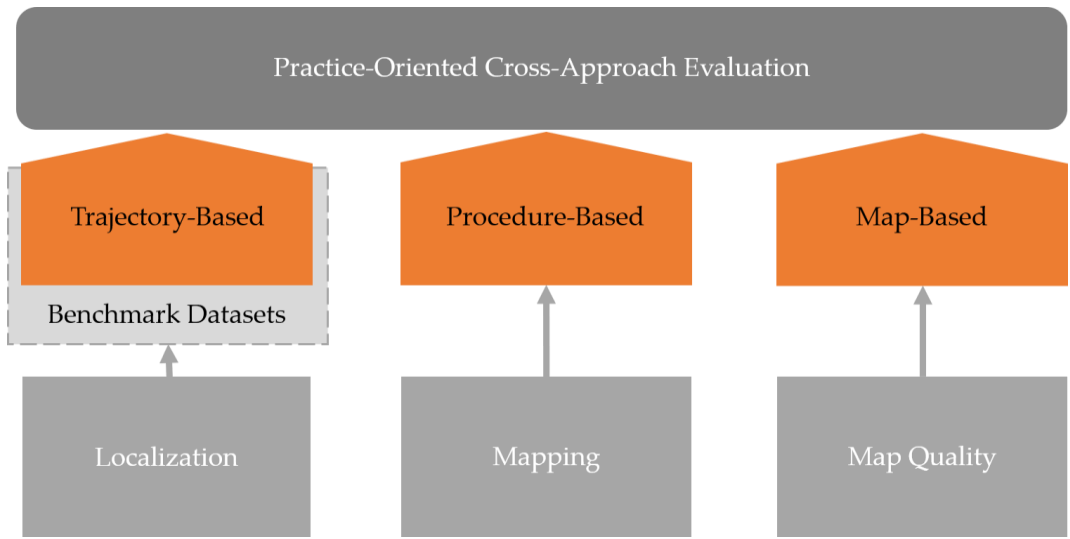

*Figure 1: Basis of the Evaluation approach*

The test scenarios are then applied to verify and evaluate the three mapping methods. For this purpose, on the one hand, a geometric verification is carried out, in which the deviation of the created maps from reality is examined. This is done by calculating the deviation between reference measurements and the corresponding measurements in the created maps. The measurements in the maps are taken using the RViz-Measuring tool [23]. Furthermore, it is qualitatively examined whether the three maps contain a complete image of the test environment.

Afterwards, the resulting maps are validated experimentally by examining their suitability for localization with AMCL (Adaptive Monte Carlo Localization). For this purpose, the previously recorded sensor data of the test scenarios are used for localization using AMCL in each map and the resulting trajectory is recorded. It is then examined by calculating the deviation between the recorded trajectory and the ground truth positions at 12 different checkpoints. Further, the completeness and correctness of the trajectory is investigated qualitatively. In order to reveal possible weaknesses of the methods, influencing factors such as different soils or environments are evaluated by examining individual areas with these characteristics. The result of the three methods is then compared.

Finally, the quality to effort ratio is calculated using the time required to execute the mapping approaches, the map accuracy as well as the localization trajectory accuracy.

## 4 EXPERIMENTAL SETUP

The setup used for conducting the experiments consists of three main aspects. First, there is the selection of the test area to conduct the evaluation of the three mapping approaches. Second, a suitable approach of each of the categories SLAM, TLS and PABC is selected. Third, the reference measurements for the defined experiments are taken.

The area that has been selected for this evaluation is part of the Hamburg University of Technology (TUHH) campus. The area is shown in Figure 2, the boundary of the area to be mapped is shown in red. The area accessible to the robot is shaded in blue and extends over a size of 220 m x 50 m. The drivable area corresponds to approximately 6,500 m² and the area to be mapped to approximately 28,000 m². The test area contains a diverse environment (see Figure 3) with different surface conditions, such as large stone pavement, small stone pavement, sandy soil and concrete stone pavement.

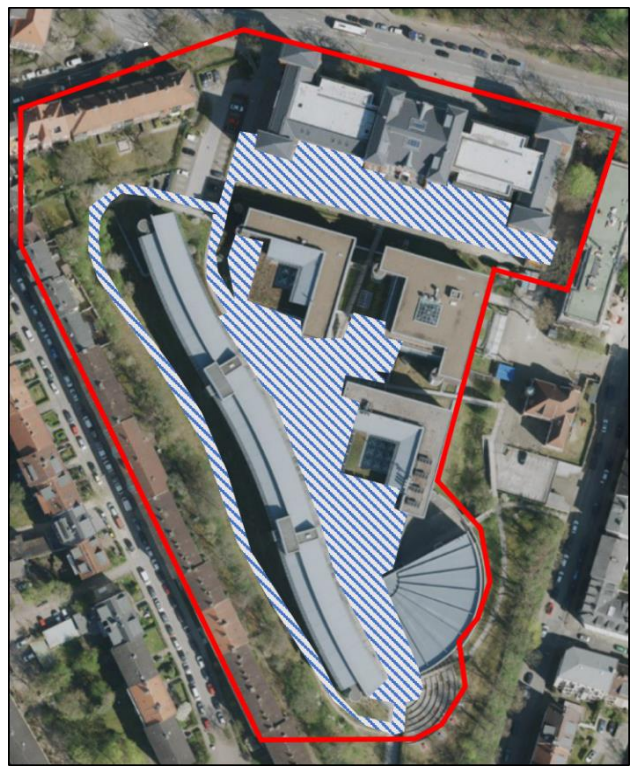

*Figure 2: Test Area at the Hamburg University of Technology*

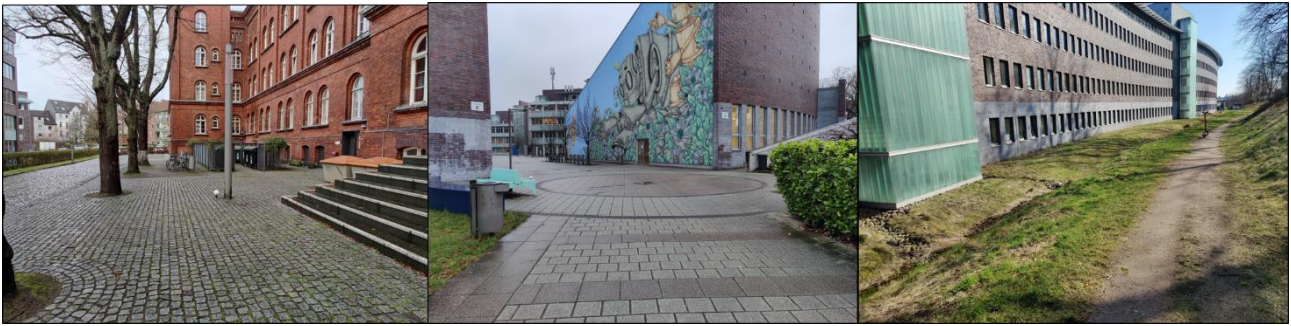

*Figure 3: Different surface conditions in the test area*

The map shows differences in height of more than 3 m and varying degrees of vegetation in the different areas. A 180 m long building which is surrounded by the drivable area is located in the middle of the test field. In addition, there is regular traffic in the form of pedestrians and cyclists, as well as irregular traffic in form of vehicles.

For the three mapping approaches a suitable algorithm and procedure are chosen. A laser SLAM package was selected due to the hardware setup of robot used for the experiments. The route for the data recording has been selected to ensure the highest possible illumination of the area and has been followed twice, once from each direction, in order to reduce possible interference from passers-by or vehicles. In addition to the mobile transport robot "Laura" developed in the scope of the TaBuLa-LOG project [24] used for recording (see Figure 4) a workstation with Ubuntu operating system and the ROS packages Google Cartographer and Rosbag are used for this purpose.

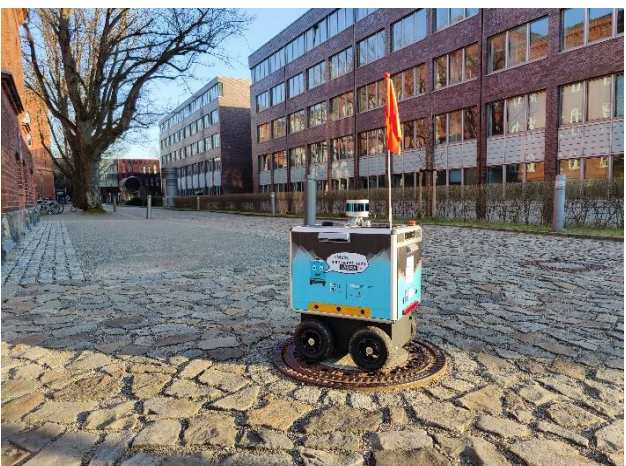

*Figure 4: Transport Robot Laura*

For TLS, a FARO Focus S70 is used in combination with the associated Faro Scene software. 23 scan and 34 sphere positions are planned at which a total of 34 full scans and 54 detail scans of spheres were taken. For the main scans a resolution of 44 million points was used and data from supporting sensor such as GPS, compass and inclinometer was captured. The acquired scans are registered and the resulting point cloud is cleaned using Faro Scene. Finally, Adobe Photoshop is used to scale the image and convert it to the required format. The TLS-setup at the TU-Campus is shown in Figure 5.

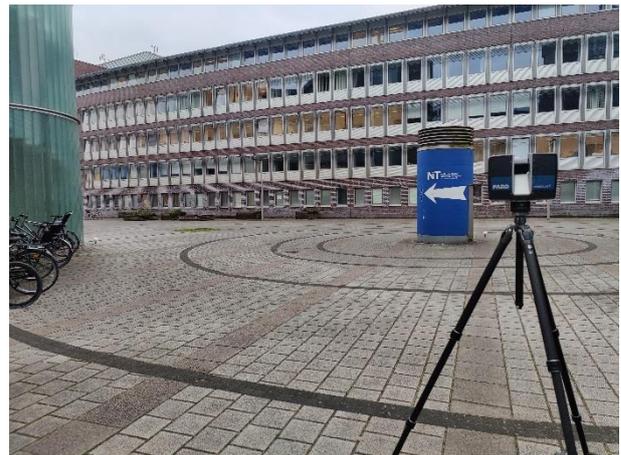

*Figure 5: TLS at TU-Campus*

OpenstreetMap is used as PABC data provider (see Figure 6). The building contours are derived using Adobe Photoshop. Then the image file is converted into the required file format and meta data is extracted using Adobe Photoshop.

The reference measurements for the quantitative evaluation of the map quality are taken using a PeakTech 2800A laser range finder mounted on a tripod. 25 reference measurements are taken across the campus as shown in Figure 7.

These include distances between buildings as well as the length of building sections to make sure that displacements or distortions can be recorded. For the evaluation of the maps, the absolute deviation is first calculated as the difference between the reference value and the digital measured value measured in the respective map. Then the average deviation is calculated over all measuring points and maxima and minima are determined.

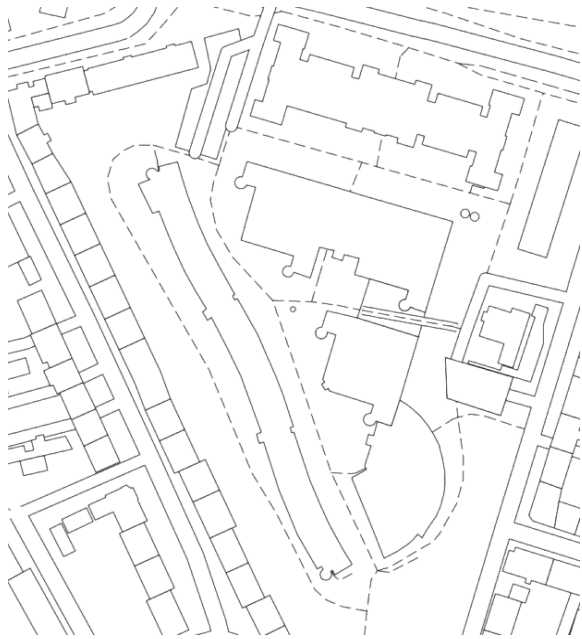

*Figure 6: OSM-Basis for PABC*

For the experimental validation of the localization, the transport robot "Laura" is used (see Figure 4). The robot is based on a Clearpath Jackal and can transport a stacking container in Euronorm format 400 x 300 mm. In addition to four Intel RealSense stereo cameras for close-range monitoring, Laura is equipped with a Velodyne VLP-16 LiDAR. This together with odometry data is used for localization with AMCL [25].

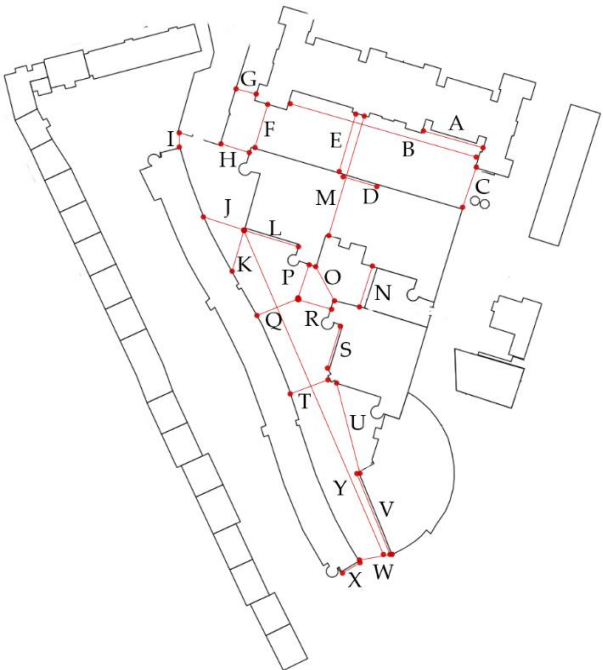

*Figure 7: Reference Measurements Geometric Verification*

For the quantitative evaluation of the localization performance, a route passing 12 checkpoints (see Figure 8) is first recorded in a Rosbag file. At each checkpoint the distances to surrounding features like walls on the map were measured using the laser range finder. The test drive was 23 min long, the corresponding Rosbag-file has a size of 7.6 GB recorded data.

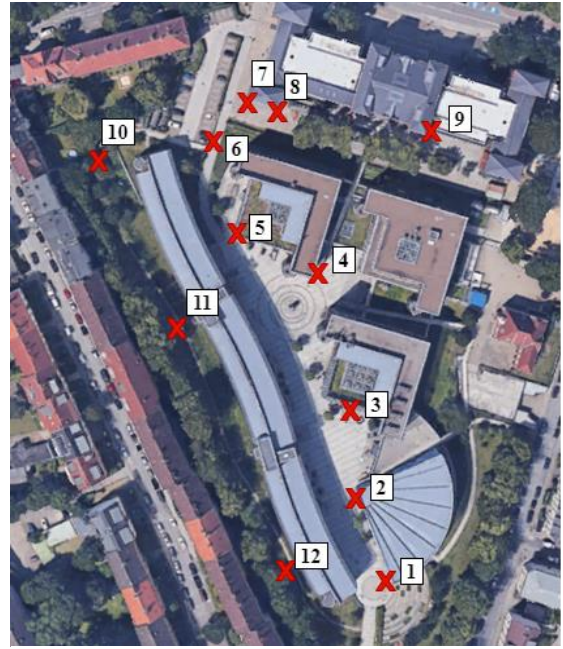

*Figure 8: Checkpoint 1-12*

Using AMCL and the respective map as a basis, the trajectory traveled during the test drive is recorded. The distance at the checkpoints between the trajectory and the features are measured using RViz. Then the deviation of all corresponding points to the reference measurement is calculated. After that, the average distance is calculated and maxima and minima are determined.

## 5 RESULTS

In this section the results of the evaluation are described and discussed. Starting with the results of the map-based evaluation, then the localization-based evaluation and lastly the procedure-based evaluation. Finally, the results will be condensed and discussed.

### 5.1 MAP-BASED EVALUATION

The results of the map-based evaluation show that all three presented methods for map generation are able to reproduce the environment true to scale. There is a big difference in the level of detail of the environment represented in the map, though. The resulting maps are shown in Figure 9. The following will use these to present the results.

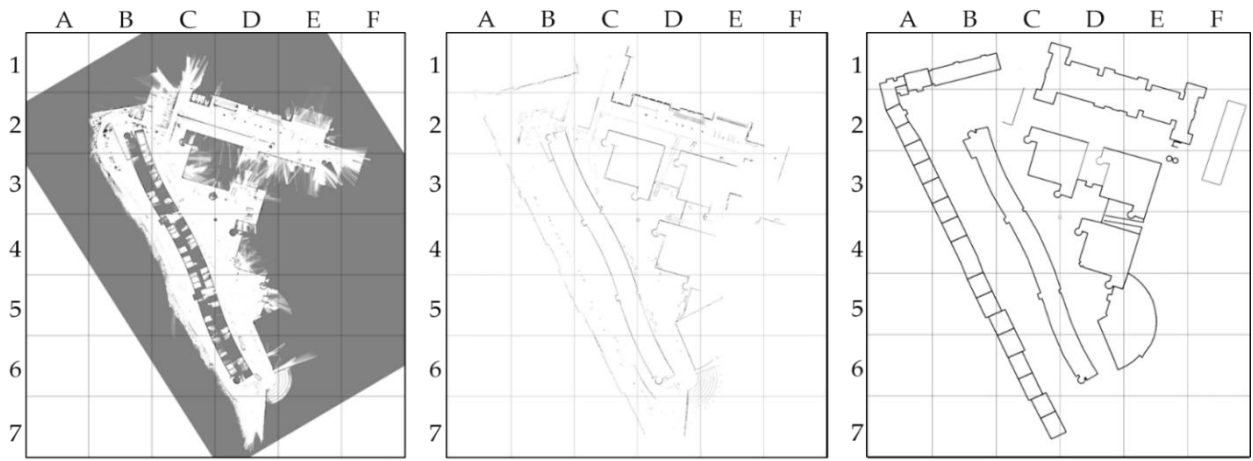

*Figure 9: Results of the three mapping approaches SLAM, TLS and PABC*

Qualitatively it is recognizable that the arrangement of buildings to each other in the maps is basically correct. However, the PABC map has a much lower level of detail. This can be seen for example in fields B2, D2 and E6. Here, details such as steps, garbage containers or vegetation are missing. The map only contains the building contours that are also visible in the source OSM map.There is also a difference in range to be seen between the SLAM map and the TLS map. Whereas the robot-based SLAM approach is only able to map the features around the actually driven passage, the TLS map also shows contours of further away buildings and obstacles. An example can be seen in D3.

The results of the quantitative evaluation of the map quality are shown in Table 1. The average and median deviation as well as the minimum and maximum deviation were determined using the 25 reference measurements recorded beforehand.

The data in Table 1 shows that TLS produces the most accurate map. With a deviation ranging from 0.05 cm to 5.60 cm averaging at 1.57 cm. The median of 1.02 cm shows that half of the measurements show a deviation of about 1 cm or less. The deviation at 21 of 25 measurements is less than 3 cm.

*Table 1: Results of Geometric Verification*

|  | SLAM | TLS | PABC |
|---|---|---|---|
| **Average** | 5.44 cm | 1.57 cm | 55.59 cm |
| **Median** | 4.39 cm | 1.02 cm | 44.25 cm |
| **MAX** | 15.54 cm | 5.60 cm | 130.13 cm |
| **MIN** | 0.67 cm | 0.05 cm | 0.60 cm |

The SLAM map, while showing an average deviation about 3.5 times as high as the one of the TLS map, still shows a respectable level of accuracy. The median is slightly lower than the average deviation, which indicates that there are highly inaccurate outlier measurements. This is confirmed by the fact, that 17 of 25 measurements show a deviation below the average of 4.39 cm.

The results for PABC confirm the impression of the qualitative analysis. The deviations are significantly higher than in the SLAM or TLS map. The average deviation of 55.59 cm is slightly more than the width of the transport robot with which the localization-based evaluation is conducted. The deviation is more than 10 times higher than in the SLAM map. The minimum deviation, on the other hand, is similar to the SLAM map, which shows that the accuracy in some areas of the map is acceptable. The maximum, on the other hand, exceeds the other two mapping methods by far.

### 5.2 Localization-Based Evaluation

The results of the localization-based evaluation show that all three presented methods for map generation are able to keep localization accurate for the majority of the drive. Figure 10 shows the trajectories derived from AMCL colored in in red (SLAM), orange (TLS) and blue (PABC). For simplicity, this figure uses the TLS map as it is the most acurate.

The qualitative examination of the trajectories shows that the localization using the TLS map and SLAM map shows mostly equivalent accuracy. Only in quadrant B4 they show a difference. Considering that the recorded journey started and ended at the same position (Checkpoint 1), it can be assumed that the localization with the help of the maps was not successful for any map towards the end. The localization confidence obviously decreases in the area of A2 and B3. The trajectory of the map created with PABC shows further inaccuracies in field B1 (see Figure 11) and B2, where it intersects with the contours of the buildings.

Due to the fact that all three trajectories have already lost localization at checkpoint 10, only checkpoints 1-10

are considered in the following quantitative evaluation. The failing localization at checkpoints 11 and 12 will be addressed at the end of this section. The results in the form of maximum and minimum deviation as well as average and median deviation are shown in Table 2.

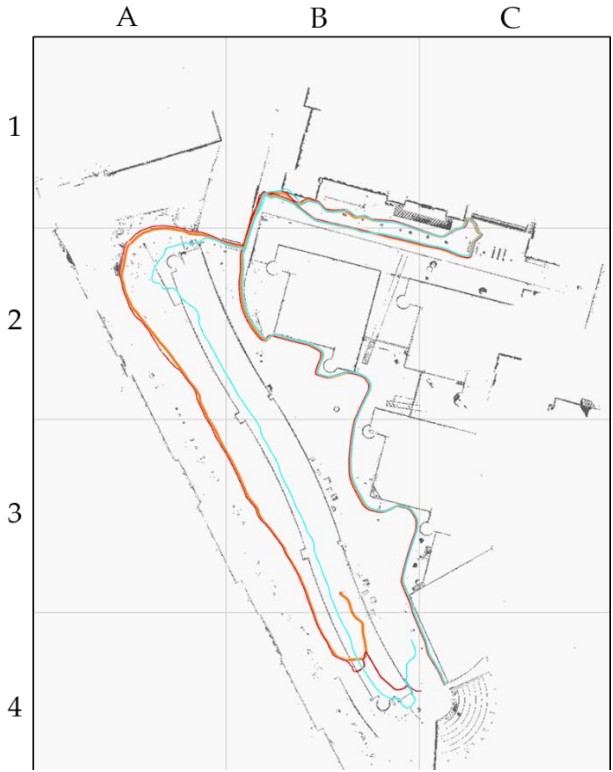

*Figure 10: Trajectories resulting from localization experiments*

The data confirms the first impression that the localization performance based on the SLAM and TLS map are comparable. Though, the TLS map achieves a slightly lower average deviation of 6.23 cm compared to the the SLAM map (7.23 cm), the minimum and maximum deviation using the SLAM map are lower than using the TLS map. However, evaluating the median localization deviations reveals that the SLAM map shows outliers with high accuracy that affect the average and that most of the values are higher than average whereas the TLS map shows the exact opposite characteristic.

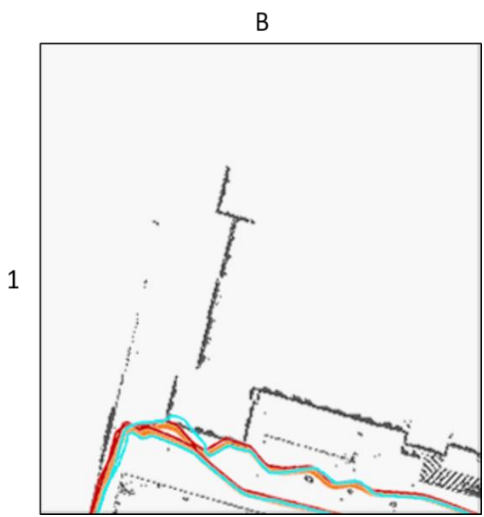

*Figure 11: Detailed view of quadrant B1 of Figure 10 showing a trajectory intersecting building contours*

*Table 2: Localization deviations in each map*

|         | SLAM     | TLS      | PABC      |
|---------|----------|----------|-----------|
| **Average** | 7.23 cm  | 6.23 cm  | 49.71 cm  |
| **Median**  | 8.83 cm  | 3.91 cm  | 36.30 cm  |
| **MAX**     | 13.56 cm | 21.50 cm | 200.98 cm |
| **MIN**     | 0.25 cm  | 0.32 cm  | 5.30 cm   |

The average deviation of the localization trajectory using the PABC map is 49.71 cm which is about 7 times the deviation of the SLAM map and about 8 times the deviation of the TLS map. In addition, the maximum deviation of SLAM and TLS is lower than the average of the PABC. The minimum deviation using the PABC map is about the average deviation of TLS. The maximum deviation of 200.98 cm with PABC testifies to a significant deviation in localization.

As mentioned in the beginning, the experiments show that the robot's localization fails in the area A2 and B3 (see Figure 10) for all three maps. Figure 12 shows photos of this locationFigure 12. An explanation for that failure is could be that the contour of the building on the one side of the path shows very few distinguishable features that might not suffice for localization. Another explanation could be the slope on the opposite side, which leads to a noisy contour who's exact position can change drastically depending on the robot's lean angle.

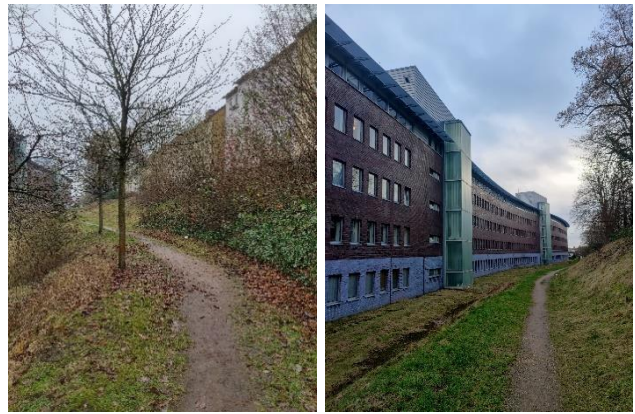

*Figure 12: Area in which localization failed*

### 5.3 PROCEDURE-BASED EVALUATION

This chapter presents the results of the evaluation of the procedure for creating the maps in the test area using SLAM, TLS and PABC. Table 3 shows the amount of data recorded as well as the time required to create the initial map (gross time) and the time required to create subsequent maps (net time) without initial setup.

*Table 3: Overview procedure-based evaluation*

|  | SLAM | TLS | PABC |
|---|---|---|---|
| **Used Data** | 14.1 GB | 246 GB | 0.0052 GB |
| **Gross Time** | 54.32 h | 90.5 h | 1.08 h |
| **Net Time** | 2.32 h | 90.5 h | 1.08 h |

Table 3 shows that the initial amount of data recorded with TLS (246 GB) far exceeds the data required for SLAM (14.1 GB) and PABC (0.0052 GB). It also shows that the procedure to create a map using TLS took significantly more time (90.5 h) than any of the other approaches. The map creation using SLAM required 54.32 h. However, a majority of this was spent in the initial setup and parameterization. Considering this only needs to be performed once, this can be reduced to a net time of 2.32 h for future map creation. PABC, on the other hand, requires the least amount of time. Since data acquisition is omitted and it is mainly a matter of deriving contours from existing map material, the time required is expectedly low.

### 5.4 DISCUSSION

Figure 13 shows the cumulated quantitative evaluation results of the procedure, localization and resulting map of each evaluated approach by putting the required effort and accuracy of the results into relation. Since there is a significant difference in gross and net time required for the SLAM approach, it is shown twice in this figure: SLAM 1 represents the evaluation results of the first map creation, SLAM 2 the ones of any subsequent map creation.

The diagram shows that SLAM with previously set up parameters performs best in comparison. Since it shows both a high map accuracy and a high localization accuracy and requires slightly over two hours (2.32 h). Including the initial effort, however, the time required is higher (54.32 h). Nevertheless the precision advantage over PABC should make up for the higher effort needed to achieve it. Overall it can be said that SLAM with the Google Cartographer is a suitable alternative for mapping, but still the localization in some areas did not work out.

The PABC results show that the required time is very low, but also the resulting map has a significantly lower feature density than the other two. The results show that it's applicability highly depends on the information provider as well as the accuracy requirements of the application. Further research effort should go into a more accurate data provider and into ways to enhance this data with, for example, aerial photography data. Generally speaking, PABC can be used in those mobile robotics applications in which a low level of details is sufficient. It is therefore a low-cost and low-effort alternative. Further, it shows high potential if better or more accurate information sources would be available.

TLS offers a highly accurate map, however, it comes with a high amount of manual effort. Only considering the quality of the results, TLS is the best candidate in this comparison. One could reduce the required effort by automating certain steps in the proceduresuch as the cleaning of the point cloud or the recording. One might achieve the former by using a height map to remove ground planes. To automate the data recording, the TLS and reference objects could be moved from one scanposition to the next by autonomous robots. With its high accuracy TLS can be considered a reasonable alternative to SLAM. A noteworthy finding is that although the precision of the TLS map is increased by a factor of 3.5, the localization is comparable.

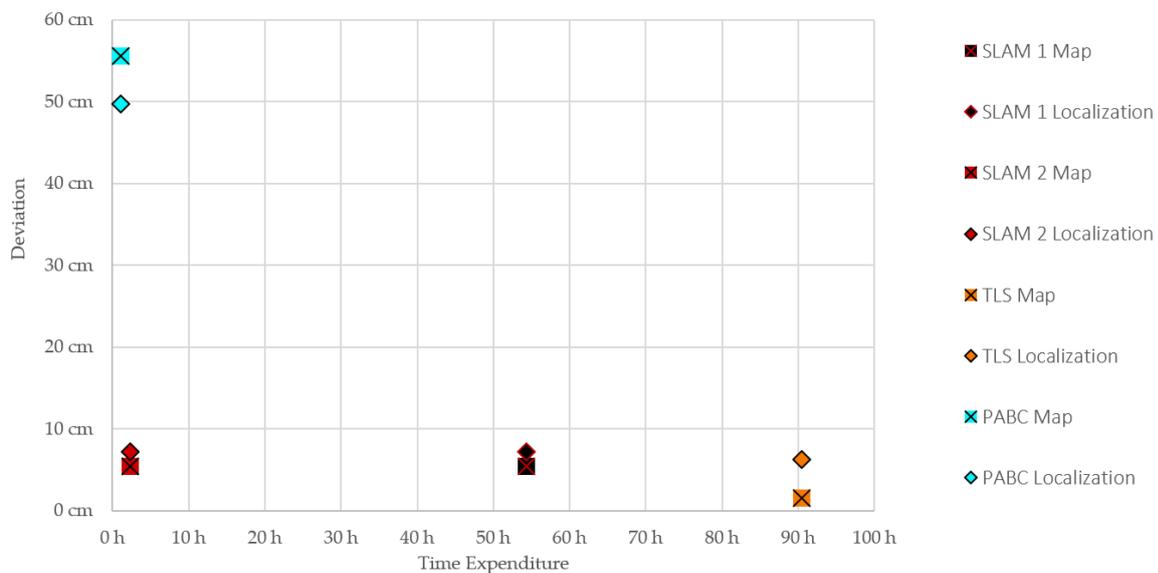

*Figure 13: Times required to create a map and their deviation from ground truth*

The evaluation conducted showed plausible results. The used metrics are suitable for identifying weaknesses of the mapping approaches. The recorded data especially in case of the robot data can be used to carry out future comparisons of other mapping approaches. The chosen test area covers different challenges of public space and is therefore a good basis for the evaluation. Though, future research should carry out this evaluation in a different area to further investigate this process. This will also yield more insights on the extent to which SLAM parametrization is transferrable to different locations.

Still, there are some aspects that need to be amended for better reliability of the results. First of all for the localization there were only 12 checkpoints used to measure the deviation, which only gives a brief overview of the localization performance. A larger amount of checkpoints or different way of capturing a ground truth trajectory could increase the expressiveness of the evaluation results. One might include a separate sensor such as a highly accurate GNSS reference to provide ground truth data.

In Figure 7 it can be seen that no reference measurements were taken on the left side of the test area. Due to the terrain on this side, no reliable measurements were possible with the given hardware equipment. More precise evaluation data in this area would have been necessary, most notably because it is the area in which the localization in all maps showed the lowest level of accuracy.

Another aspect which could be improved is the usage of the RViz-measuring tool. It calculates the distance between two points marked manually using the pixel distance and the resolution. Using a suitable automated measuring method would lead to less inaccuracies and less required manual effort.

Finally, the time expenditure that has been recorded during the experiments is subject to the user's level of experience with the respective system. It should be considered a guideline for inexperienced users rather than an absolute value.

## 6 Conclusion

In this paper, three different mapping approaches have been evaluated and compared with regards to required hardware and effort as well as the overall quality of the resulting maps and their applicability for real-live applications. The compared mapping approaches are based on simultaneous localization and mapping (SLAM), terrestrial laser scanning (TLS) and publicly available building contours (PABC). The specific implementations of these approaches that were evaluated were the Google Cartographer [26], a Faro Focus S70 with Faro Scene [27], and building contours provided by OpenStreetMap [28].

For the comparison a 28.000 m² area at the campus of the Hamburg University of Technology was selected and mapped with each of the mapping approaches. Ground truth data of buildings and previously selected checkpoints was gathered using a laser distance meter. This data has been used for the first step of the evaluation, a quantitative analysis of the resulting maps. In the second step a mobile robot was driven along multiple, predefined routes across the selected area and past the aforementioned checkpoints. The collected sensor data was then used to retrace the driven path in each created map using Adaptive Monte-Carlo Localization [25, 29]. The resulting trajectories were evaluated using the preselected checkpoints.

The results show that all three of the employed approaches can be used to generate maps that are suitable for the localization and navigation of mobile robots. The TLS-based approach required the most effort in the mapping process but also showed with a deviation of 0.05 cm to 5.60 cm the highest level of accuracy in the resulting maps. The localization inaccuracy ranged from 0.32 cm to 21.50 cm. The PABC-based approach required the least amount of effort and yielded maps with the lowest level of detail. Depending on the granularity of the provided building contours, this approach showed a deviation in building contours of 0.60 to up to 130.13 cm and in localization of 5.30 to 200.98 cm. Lastly, the SLAM-based approach took more effort than the PABC-based one but still considerably less than the TLS-based approach while at the same time yielding results comparable to the ones created by the TLS-based approach. The building contours deviated between 0.67 cm and 15.54 cm from the ground truth, the localization inaccuracy by 0.25 to 13.56 cm.

Apart from the actual evaluation results, this work has yielded three main insights: 1) When selecting the mapping approach for a specific robotic application it is worth looking beyond the best-known candidates. Other approaches might be more suitable with respect to available hardware or ground conditions in the respective area while still yielding sufficiently accurate results. 2) Despite their low level of detail, PABC-based maps can be applicable for usage in mobile robotics applications and are a low-cost and low-effort alternative to more established approaches such as SLAM or TLS. However, it is not suited for applications in which a precise localization is required or the provided satellite data is highly inaccurate. 3) Quantitative measures of map quality do generally correlate to a map's applicability for the localization of mobile robots. However, at a certain level of precision the gain in localization accuracy becomes negligible.

FINANCIAL DISCLOSURE

This article is based on research that was undertaken in the project "TaBuLa-LOG - Kombinierter Personen- und Warentransport in automatisierten Shuttles" sponsored by the German Federal Ministry of Transport and Digital Infrastructure between 2020 and 2021.

**Justin Ziegenbein, M.Sc.,** Research Assistant at the Institute for Technical Logistics, Hamburg University of Technology. Justin Ziegenbein studied Logistics and Mobility and International Management and Engineering at Hamburg University of Technology.

**Manuel Schrick, M.Sc.,** Research Assistant at the Institute for Technical Logistics, Hamburg University of Technology. Manuel Schrick studied Computer Science at University of Lübeck and RWTH Aachen University.

**Marko Thiel, M.Sc.,** Research Assistant at the Institute for Technical Logistics, Hamburg University of Technology. Marko Thiel studied Mechanical Engineering and Theoretical Mechanical Engineering at Hamburg University of Technology.

**Dr. Johannes Hinckeldeyn**, Senior engineer at the Institute for Technical Logistics, Hamburg University of Technology. After completing his doctorate in Great Britain, Johannes Hinckeldeyn worked as Chief Operating Officer for a manufacturer of measurement and laboratory technology for battery research. Johannes Hinckeldeyn studied industrial engineering, production technology, and management in Hamburg and Münster.

**Prof. Dr.-Ing. Jochen Kreutzfeldt**, Professor and Head of the Institute for Technical Logistics, Hamburg University of Technology. After studying mechanical engineering, Jochen Kreutzfeldt held various managerial positions at a company group specializing in automotive safety technology. Jochen Kreutzfeldt then took on a professorship for logistics at the Hamburg University of Applied Sciences and became head of the Institute for Product and Production Management

Address: Institute for Technical Logistics, Hamburg University of Technology, Theodor-Yorck-Strasse 8, 21079 Hamburg, Germany; Phone: +49 40 42878-4861, E-Mail: justin.ziegenbein@tuhh.de